\documentclass[letterpaper, 10 pt, conference]{ieeeconf}  % Comment this line out if you need a4paper

\usepackage{microtype}
\usepackage{hyperref}
\usepackage{url}
\usepackage{booktabs}
\usepackage{multirow}
\usepackage{graphicx}
\let\labelindent\relax
\usepackage{enumitem}
\usepackage[most]{tcolorbox}
\usepackage{wrapfig}
\usepackage{floatflt}
\usepackage{color, colortbl}
\usepackage{listings}
\makeatletter
\let\old@lstKV@SwitchCases\lstKV@SwitchCases
\def\lstKV@SwitchCases#1#2#3{}
\makeatother
\usepackage{lstlinebgrd}
\makeatletter
\let\lstKV@SwitchCases\old@lstKV@SwitchCases

\lst@Key{numbers}{none}{%
    \def\lst@PlaceNumber{\lst@linebgrd}%
    \lstKV@SwitchCases{#1}%
    {none:\\%
     left:\def\lst@PlaceNumber{\llap{\normalfont
                \lst@numberstyle{\thelstnumber}\kern\lst@numbersep}\lst@linebgrd}\\%
     right:\def\lst@PlaceNumber{\rlap{\normalfont
                \kern\linewidth \kern\lst@numbersep
                \lst@numberstyle{\thelstnumber}}\lst@linebgrd}%
    }{\PackageError{Listings}{Numbers #1 unknown}\@ehc}}
\makeatother

\definecolor{mybg}{rgb}{0.95, 0.95, 0.95}
\tcbuselibrary{listingsutf8}
\lstdefinelanguage{LTL}{
  morekeywords={LTLSPEC,NAME,G,F,X,U,R,W,TRUE,FALSE},
  sensitive=true,
  morecomment=[l]{//},
  morestring=[b]",
}
\lstdefinelanguage{NuSMV}{
  morekeywords={MODULE,VAR,ASSIGN,init,next,case,esac,TRUE,FALSE,boolean},
  sensitive=true,
  morecomment=[l]{--},
  morestring=[b]",
}
\lstdefinestyle{mylststyle}{
  basicstyle=\ttfamily\small,
  keywordstyle=\color{RoyalBlue}\bfseries,
  commentstyle=\color{green!50!black}\itshape,
  stringstyle=\color{violet},
  showstringspaces=false,
  tabsize=2,
  breaklines=true,  
  columns=fullflexible, 
  keepspaces=true,
}
\usepackage{tikz}
\usetikzlibrary{tikzmark}

\tcbset{
  myboxstyle/.style={
    boxrule = 0pt,
    toprule = 4.5pt, % top rule weight
    enhanced,
    fuzzy shadow = {0pt}{-2pt}{-0.5pt}{0.5pt}{black!35}, % {xshift}{yshift}{offset}{step}{options}
    fontupper = \ttfamily\small, % Use a monospaced font (e.g., Courier)
    boxsep = 5pt, % padding inside the box
    left = 1pt, % left padding
    right = 1pt, % right padding
    top = 5pt, % top padding
    bottom = 5pt, % bottom padding
  }
}

\usepackage{amsthm}
\usepackage{amssymb}
\usepackage{mathtools}
\usepackage[dvipsnames]{xcolor}  % Load color names like RoyalBlue
\usepackage{hyperref}
\usepackage[nameinlink,capitalize]{cleveref}
\hypersetup{colorlinks=true,linkcolor=RoyalBlue,citecolor=RoyalBlue,urlcolor=RoyalBlue,pdfborder={0 0 0}}
\usepackage[normalem]{ulem} % for command \sout "strike out"
\usepackage{mathtools}

\usepackage[utf8]{inputenc} % allow utf-8 input
\usepackage[T1]{fontenc}    % use 8-bit T1 fonts
\usepackage{hyperref}       % hyperlinks
\usepackage{url}            % simple URL typesetting
\usepackage{booktabs}       % professional-quality tables
\usepackage{amsfonts}       % blackboard math symbols
\usepackage{nicefrac}       % compact symbols for 1/2, etc.
\usepackage{microtype}      % microtypography
\usepackage{color, colortbl}
\definecolor{Gray}{gray}{0.9}
\newcolumntype{g}{>{\columncolor{Gray}}c}

\usepackage{graphicx,wrapfig}
\usepackage{algorithm}
\usepackage[noend]{algpseudocode}
\usepackage{caption}
\usepackage{subcaption}
\usepackage{url}
\theoremstyle{definition}

\theoremstyle{remark}

\newcommand{\bc}{\begin{center}}
\newcommand{\ec}{\end{center}}

\newcommand{\bdm}{\begin{displaymath}}
\newcommand{\edm}{\end{displaymath}}

\newcommand{\beq}{\begin{equation}}
\newcommand{\eeq}{\end{equation}}

\newcommand{\bfl}{\begin{flushleft}}
\newcommand{\efl}{\end{flushleft}}

\newcommand{\bt}{\begin{tabbing}}
\newcommand{\et}{\end{tabbing}}

\newcommand{\beqn}{\begin{align}}
\newcommand{\eeqn}{\end{align}}

\newcommand{\beqs}{\begin{align*}} % no equation numbers
\newcommand{\eeqs}{\end{align*}}  % no equation numbers

\definecolor{Highlight}{rgb}{0.92,0.94,1}

\newcommand{\OURS}{LAD-VF}

\IEEEoverridecommandlockouts                              
\title{\LARGE \bf
\OURS: LLM-Automatic Differentiation Enables \\ Fine-Tuning-Free Robot Planning from Formal Methods Feedback
}

\author{Yunhao Yang$^{1}$, Junyuan Hong$^{1}$, Gabriel Jacob Perin$^{2}$, Zhiwen Fan$^{3}$, Li Yin$^{4}$, Zhangyang Wang$^{1}$, Ufuk Topcu$^{1}$
\thanks{$^{1}$
The University of Texas at Austin, Austin, TX, United States;$^{2}$
University of São Paulo, São Paulo, SP, Brazil;
$^{3}$ Texas A\&M University, College Station, TX, United States 
$^{4}$ SylphAI, TX, United States 
}
}

\begin{document}

\maketitle
\thispagestyle{empty}
\pagestyle{empty}

\begin{abstract}
% Large language models (LLMs) offer powerful and flexible interfaces for controlling complex systems, yet aligning their outputs with domain-specific formal specifications remains challenging. Existing methods rely heavily on enormous human feedback or costly fine-tuning, limiting scalability and interpretability. To overcome these limitations, we propose \textbf{\OURS}, a fine-tuning-free framework that integrates automatic prompt engineering with formal verification feedback to enable LLMs to comply with formal specifications in a few shots. Specifically, we transform LLM-based pipelines into differentiable computation graphs, leveraging formal verification outcomes as structured supervision signals. By treating prompts and textual inputs as trainable parameters, our approach systematically refines model behaviors to satisfy domain constraints. 

Large language models (LLMs) can translate natural language instructions into executable action plans for robotics, autonomous driving, and other domains. Yet, deploying LLM-driven planning in the physical world demands strict adherence to safety and regulatory constraints, which current models often violate due to hallucination or weak alignment. Traditional data-driven alignment methods, such as Direct Preference Optimization (DPO), require costly human labeling, while recent formal-feedback approaches still depend on resource-intensive fine-tuning.
In this paper, we propose \OURS, a fine-tuning-free framework that leverages formal verification feedback for automated prompt engineering. By introducing a formal-verification-informed text loss integrated with LLM-AutoDiff, \OURS\ iteratively refines prompts rather than model parameters. This yields \textbf{three key benefits}: \textbf{(i)} scalable adaptation without fine-tuning;  \textbf{(ii)} compatibility with modular LLM architectures;  and \textbf{(iii)} interpretable refinement via auditable prompts.
Experiments in robot navigation and manipulation tasks demonstrate that \textbf{\OURS} substantially enhances specification compliance, improving success rates from 60\% to over 90\%. Our method thus presents a scalable and interpretable pathway toward trustworthy, formally-verified LLM-driven control systems.

% while reducing the end-to-end inference time by 37.9\%. Code will be available.

\end{abstract}

\section{Introduction}

Large language models (LLMs)~\cite{brown2020language} have revolutionized high-level decision making in domains such as robotics~\cite{Song2022LLMPlannerFG, Liu2023LLMPEL}, autonomous driving~\cite{yao2022react}, and software verification~\cite{yang-mlsys}. LLMs enable the translation from task instructions in natural language to action plans that are executable by machines, offering a flexible and general-purpose interface for downstream tasks~\cite{CodeBotler, Singh2022ProgPromptGS}. 
However, adapting the LLM-based method in the physical world faces a major challenge in safety: Running a robot in the physical world should not only achieve the goal, e.g., driving toward the specified spot, but also have to comply with physical constraints (e.g., safety specifications) or societal regulations (e.g., traffic rules).
Yet, existing LLMs suffer from hallucination or are not well aligned for generating constrained action plans, leaving safe LLM-driven action planning as an open challenge~\cite{wang2024llm, yang2023planning, chen2024can}. 

Data-driven alignment is a plausible solution that utilizes human feedback~\cite{dpo, achiam2023gpt, suzgun2022challenging} to reward responses against undesired and maximizes the chance for LLMs to generate desired outputs.
For example, Direct Preference Optimization (DPO) \cite{dpo} contradicts the pair of preferred and rejected responses in training.
Such a method, however, is labor-intensive and difficult to scale up without sufficient human labeling of the preference data.

\begin{figure*}[t]
    \centering
    \includegraphics[width=0.95\linewidth]{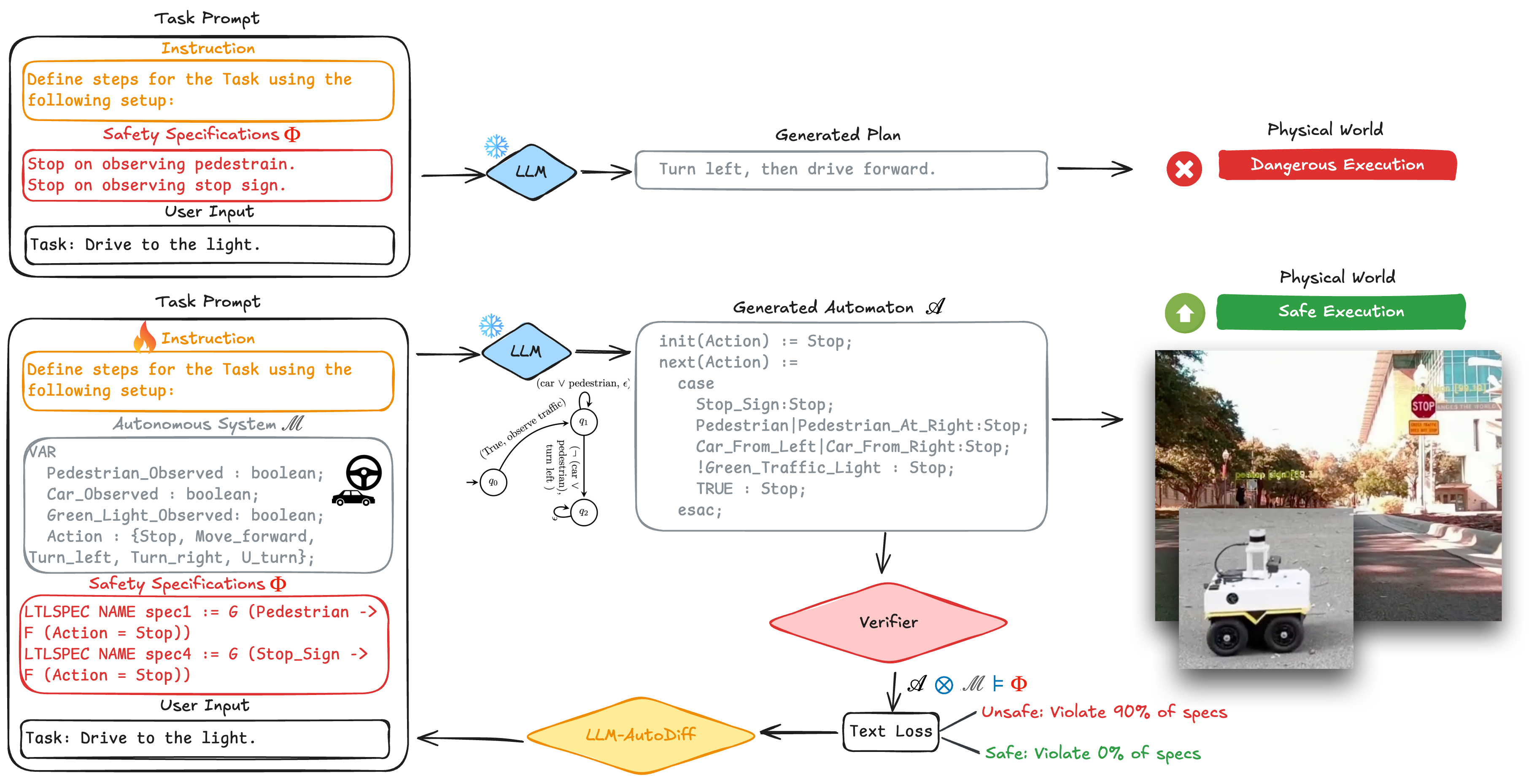}
    \caption{The diagram illustrates a closed-loop planning framework for generating and verifying plans for autonomous systems. The user provides a task prompt, which the language model uses to generate a plan. The plan is converted into an automaton $\mathcal{A}$ and verified against safety specifications $\Phi$ when operating in the system $\mathcal{M}$. The number $n_f$ of failed specifications is a loss for \OURS \ to optimize the task prompt iteratively.}
    \label{fig: safe-pipeline}
    \vspace{-0.2in}
\end{figure*}

Recently, leveraging \emph{formal methods feedback} for automatic preference labeling emerged as a promising alternative to the traditional DPO~\cite{yang-mlsys}.
Yang et al. proposed using formal methods, such as model checking, to provide feedback for fine-tuning LLMs, enabling them to generate high-fidelity planning solutions that comply with formal rules. 
In particular, this work transforms LLM outputs into finite-state automata and formally verifies the automata against pre-defined logical specifications. Then, it treats the number of specifications satisfied by each automaton as the ``reward'' for fine-tuning.
Yet, existing methods that enable fine-tuning using formal methods feedback \cite{yang-mlsys, yang2024joint} demand more data, and decisions are generated from a black box. 
While the methods eliminate the need for human labels, they typically require massive training data and computational resources for loss convergence, limiting their scalability to larger models.

In this work, we propose a fine-tuning-free method, \textbf{L}LM-\textbf{A}uto\textbf{D}iff from \textbf{V}erification \textbf{F}eedback or \textbf{\OURS}, that improves the safety compliance via automatic prompt engineering~\cite{zhou2022large} instead of fine-tuning model parameters.
Modern LLMs with large-scale pre-training can be easily steered via a proper textual prompt, including task instructions and essential context~\cite{dong2022survey, white2023prompt, marvin2023prompt}.
As demonstrated in \cref{fig: safe-pipeline}, we streamline the feedback from the verifier to improve the LLM behaviors via automatically updating the prompts.
Given the safety feedback, we leverage LLMs to generate textual improvement on textual prompts, which were formulated as LLM-Automatic Differentiation or LLM-AutoDiff~\cite{yin2025llm-autodiff}.

Our major technical contribution lies in a novel \textbf{formal-verification-informed text loss} integrated with the LLM-AutoDiff, enabling automated prompt engineering from formal feedback. 
Our approach has several unique advantages compared to traditional ones:
\begin{itemize}
    % provide structured, reliable supervision that is domain-aligned yet automatable, avoiding reliance on noisy human preference data.
     \item \textbf{Fine-tuning-free adaptation:} eliminates the need for costly parameter updates, enabling scalable use of LLMs in new safety-critical domains.
    \item \textbf{Compatibility with modular LLM architectures:} adapts to varied modular LLM pipelines, and removes the need for parameter fine-tuning when the query format or component structure changes.  
    \item \textbf{Improved interpretability:} enables transparent and auditable changes by refining prompts rather than hidden weights, making the improvement explainable.
\end{itemize}
We validate our approach in settings where autonomous agents must follow natural language instructions while complying with formal task constraints, such as autonomous driving within rulebooks or robotic manipulation governed by safety logic. \textbf{\OURS} substantially improves compliance with formal specifications---boosting success rates from 60\% to over 90\%---while maintaining generalization and interpretability. More broadly, our framework unifies automated prompt optimization with formal verification feedback, offering a scalable path toward \emph{trustworthy and verifiable LLM-based control systems}.

% \textcolor{red}{Below are not our innovation but AutoDiff's. But if we can demonstrate these designs are important for the specification generation, we can try to include them. @Yunhao, do we have ablation study on these strategies?}
% \begin{itemize}
%     \item \textbf{Pass-through gradients} ensure that feedback can traverse non-trainable components such as simulators or spec-checkers, informing earlier prompts that led to failure.
%     \item \textbf{Time-sequential feedback} aligns specification violations with the correct invocation of cyclic or reentrant LLM components, a crucial feature for controllers in iterative or agent-based architectures.
%     \item \textbf{Selective gradient updates} concentrate optimization effort on outputs that violate specifications, enabling efficient, spec-driven learning.
% \end{itemize}

\vspace{-0.1in}
\section{Related Works}

\textbf{Learning from Human Feedback} is a well-developed approach for enhancing foundation models. Reinforcement Learning from Human Feedback (RLHF) uses human preferences to train a reward model, which in turn guides the fine-tuning of the language model \cite{RLHF, Ouyang2022FollowInstructions}. Then, methods such as Direct Preference Optimization (DPO) streamline this process by directly optimizing the model against preference comparisons, avoiding the need for an explicit reward model \cite{Rafailov2023DPO}. These methods have shown strong empirical performance across various language tasks. However, they are labor-intensive due to the need for human annotations, and the reliance on subjective feedback makes them incapable of safety-critical tasks.

\textbf{Learning from Formal Feedback} is an alternative to human feedback, where system requirements are encoded as structured specifications such as temporal logic formulas or checklists \cite{yang-mlsys, CodeBotler}. The generated outputs can then be verified against these specifications using mathematical tools such as model checker \cite{yang2024aamas}. Existing works have demonstrated that verification outcomes can be used as feedback for refining models, either by treating the satisfaction rate of specifications as reward value or by ranking these rates \cite{yang-mlsys, bhattknow}. 
While such methods eliminate the need for human labels, they typically require large amounts of training data and computational resources to converge. In contrast, we focus on optimizing input prompts rather than model parameters via formal feedback, alleviating the need for computational resources and human labels.

\textbf{Automatic Prompt Engineering (APE)} has rapidly evolved into a diverse line of research aimed at systematically improving prompts for large language models. The seminal APE framework by \cite{zhou2022large} pioneered the idea of iteratively refining prompts through paraphrasing and selection, laying the foundation for treating prompt design as an optimization problem. Building on this, subsequent approaches explored different optimization paradigms: DLN1 \cite{sordoni2023joint} and OPRO \cite{yang2024large} framed prompt learning as distributional optimization or iterative refinement with task demonstrations; TextGrad \cite{yuksekgonul2024textgrad} introduced the notion of interpreting textual feedback as gradients to guide descent; DSPy \cite{khattab2024dspy} formalized structured prompt optimization with modular optimizers like COPRO; and PromptAgent \cite{wangpromptagent} extended the paradigm toward agent-based planning with search strategies. Meanwhile, ProTeGi \cite{pryzant2023automatic} was among the first to explicitly incorporate gradient descent principles into automatic prompt generation. 
More recently, LLM-AutoDiff~\cite{yin2025llm-autodiff} was proposed to generalize such frameworks to more complicated AI-based applications, handling cyclic computation graph. Collectively, these methods highlight the growing recognition of APE as a principled framework for automating prompt design and, therefore, enable the self-evolving of AI agents to outperform reinforcement learning~\cite{agrawal2025gepa}.
Yet, the existing self-improving prompting methods are only applied in scenarios where the correct answers are given independently of the LLM outputs (actions).
In this paper, we focus on improving LLM prompts based on dynamic verifications that depend on the actions predicted by the prompted LLM.

% , advancing from evolutionary strategies to gradient-inspired approaches and structured optimization.

% \input{sec/preliminary}
\section{Safety-Constrained Robot Planning}

In this section, we introduce \textbf{\OURS} in robot planning applications where safety rules are enforced. We query the LLM to generate robot-executable plans and verify the plans against user-provided safety specifications expressed in temporal logic formulas \cite{Clarke2018modelchecking}. We apply \OURS \ to improve the generated plans by raising the number of safety specifications being satisfied by those plans.

\begin{figure}[t]
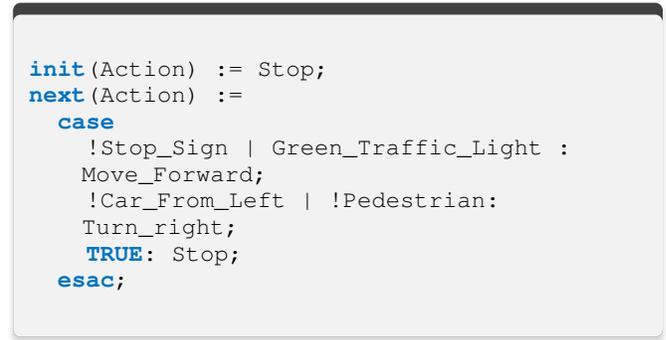

    \centering
\begin{tcolorbox}[
  % title={\small LLM-generated actions are highlighted},
  % left=2pt, right=2pt, bottom=1pt, top=1pt,
  myboxstyle
]
\begin{lstlisting}[
  language=NuSMV,
  style=mylststyle,
]
init(Action) := Stop;
next(Action) :=
  case    
    !Stop_Sign | Green_Traffic_Light : Move_Forward;
    !Car_From_Left | !Pedestrian: Turn_right;
    TRUE: Stop;
  esac;
\end{lstlisting}
\end{tcolorbox}
    \caption{An example of a NuSMV-based plan.}
    \vspace{-15pt}
    \label{fig: plan-example}
\end{figure}

\paragraph{Pipeline Overview}
Outlined in \cref{fig: safe-pipeline}, We design a pipeline following \cite{yang2024aamas} that queries an LLM to generate formally verifiable plans for robotic tasks and applies \OURS \ to optimize the plan based on the verification outcome. In particular, we first send a natural language task description (e.g., go straight at the traffic light intersection) to an LLM and extract a plan in NuSMV \cite{Cimatti2002NuSMV}---a logic-based formal language. We show an example of a NuSMV-based plan in Fig. \ref{fig: plan-example}.
We generate a plan by
\begin{align}
    \text{PlanGen}(T, \mathcal{P}) = \text{LLM}(\pi_{\text{plan}}(T, \mathcal{P}))
\end{align}
where $T$ is the task description and $\mathcal{P}$ is the set of prompts to be optimized.
$\pi_{\text{plan}}$ represents a template for the inference with $T$ and $\mathcal{P}$.

Next, we provide a set of safety specifications in logic formulas and apply a model checker \cite{Cimatti2002NuSMV} to mathematically prove whether the generated plan satisfies the specifications. Then, we record the percentage of specifications being violated and use this percentage as a feedback signal (e.g., loss) to \OURS, which will eventually optimize the plans to minimize the percentage of specification violation.

\paragraph{Automatic Prompt Engineering via LLM-AutoDiff}
A core challenge in our method is how to optimize prompts without resorting to costly fine-tuning. We want a fine-tuning-free approach that can adapt prompts at test time while remaining transparent and interpretable. Automatic Prompt Engineering (APE)~\cite{zhou2022large} offers such a solution by automating the refinement of prompts instead of altering model weights. It employs a two-engine setup: a ``\emph{forward}'' LLM performs the task, while a frozen ``\emph{backward}'' LLM critiques the outputs and proposes edits. These critiques, known as textual gradients, function like gradient descent in neural networks—providing systematic feedback in natural language to iteratively update prompts, which also enhances the explainability of the optimization. 
% For example, if a retrieval step consistently misses relevant documents, the backward engine can suggest rewording to emphasize coverage, leading to explainable, fine-tuning-free improvements over time.

However, APE alone becomes insufficient in our setting, where the pipeline must combine functional modules (e.g., a formal-method verifier) and sequential multi-step planning. Standard APE methods focus on optimizing single prompts, but they cannot propagate gradients through non-LLM components or preserve temporal order when prompts are invoked across multiple planning steps. LLM-AutoDiff~\cite{yuksekgonul2024textgrad,yin2025llm-autodiff} emerges as a unified framework that can back-propagate textual gradients (feedback) through a complex network. TextGrad first proposed a general textual gradient framework~\cite{yuksekgonul2024textgrad}. Later, \textbf{Adalflow}~\cite{yin2025llm-autodiff} further closes this gap by treating the entire workflow as a differentiable graph: pass-through gradients allow feedback from functional verifiers to influence upstream prompts, time-sequential gradients ensure that each stage in a multi-step plan is updated in order, and selective gradient computation reduces overhead by focusing only on failed examples. This unified approach makes LLM-AutoDiff a natural fit for our verifier-augmented, sequential system, enabling scalable and efficient optimization where manual prompt engineering or single-node APE would fall short.

% \subsection{\OURS Method}

\OURS\ extends Automatic Prompt Engineering into a fully auto-differentiable framework for optimizing complex LLM pipelines. Formally, we model the system as a directed graph $G = (N, E)$, where each node $v \in N$ can be an LLM module (with trainable prompt $P_v$) or a functional module. Given a set of tasks $\mathcal{T}$, the system aims to minimize a loss over the set of prompts $\mathcal{P} = \{P_v | v\in N\}$:  
\begin{align}
\mathcal{P}^{*} = \arg\min_{\mathcal{P}} \bigcup_{T\in \mathcal{T}} \mathcal{L}\!\big(\text{PlanGen}(T, \mathcal{P})\big).
\end{align} 
This formulation ensures that both LLM prompts and upstream dependencies of functional nodes can be optimized under a unified objective.  
During training, a forward pass executes all nodes in topological order, while a backward pass propagates \emph{textual gradients} -- the feedback on how to update the prompt/intermediate outputs. For an internal node $v$, the gradient is aggregated from its successors $w$:  
\begin{align}
\frac{\partial \mathcal{L}}{\partial v} = 
\bigcup_{w \in \text{SuccessorsOf}(v)} 
\text{LLM}_{\text{backward}}\!\left(v, w, \frac{\partial \mathcal{L}}{\partial w}\right),
\end{align}
where $\text{LLM}_{\text{backward}}$ represents a backward inference generating the feedback by LLMs.

By default, we adopt the Adalflow in our framework for the below two reasons:
(1) Functional nodes (e.g., verifier modules) have no prompts to update, but Adalflow introduces \emph{pass-through gradients} so their outputs still propagate error signals upstream, allowing verifier feedback to refine earlier prompts.  
(2) For sequential prompting, where the same node is invoked multiple times in a plan, Adalflow attaches timestamps $t$ to each call, yielding \emph{time-sequential gradients}, which ensures that updates respect the chronological order of multi-step plans.  
% (3) Finally, to reduce computation, \OURS applies \emph{selective gradient computation}, only generating backward signals for failed examples, avoiding redundant updates on already correct outputs. 
Prompt updates are then synthesized by an optimizer LLM:  
\begin{equation}
\mathcal{P}_v^{\text{new}} 
= \text{LLM}_{\text{opt}}\!\Big(
\mathcal{P}_v, \; \text{GradientContext}(v), \; 
\frac{\partial \mathcal{L}}{\partial v}\Big).
\end{equation}
Together, we are able to update prompts without fine-tuning.

\paragraph{Formal Verification as Textual Feedback}
A central step of the framework is to extract a loss that guides prompt optimization from formal verification outcomes. We first query the LLM to convert the generated plan into an automaton, expressed in NuSMV. This conversion enables the generated plan to be checked against a set of logical specifications provided by the user.

Once the automaton is generated, we apply a model checker to verify whether it satisfies the given specifications. Each specification returns a binary signal (satisfied or violated). To convert these outcomes into a quantitative supervision signal, we define the formal feedback loss as $\mathcal{L} = n_f / n_{\text{total}}$,
where $n_f$ is the number of violated specifications and $n_{\text{total}}$ is the total number of specifications provided. 
% However, the $\mathcal{L}_{\text{formal}}$ is not differentiable since it is not connected to the text parameter (system prompt) directly in the computation graph.
% Instead, we provide a textual loss that can be backward.
% The textual loss is a formatted text $\mathcal{L} = $``\texttt{Task: \{\{task\}\}$\backslash$nSolution: \{\{plan\}\}$\backslash$nScore: \{\{$\mathcal{L}_{\text{formal}}$\}\}}''.

% $\mathcal{L}_{\text{spec}}$ is then integrated into the overall optimization objective $\mathcal{L}$:
% \[
% \mathcal{L} \;=\; \mathcal{L}_{\text{spec}} \;+\; \mathcal{L}_{\text{other}},
% \]
% where $\mathcal{L}_{\text{other}}$ captures additional losses arising from other components of the pipeline.
% If a non-safety relevant task objective exists and is evaluable, $\mathcal{L}_{\text{other}}$ can be used to capture the objective completion rate. For example, $\mathcal{L}_{\text{other}} = 0 \text{ or } 1$ depends on whether the robot has reached the target destination.
% In our experimental settings, where safety is the only objective, $\mathcal{L}_{\text{other}}$ is always equal to 0.

For example, if a generated plan is verified against $15$ safety specifications and $3$ of them are violated, the resulting loss is $\mathcal{L} = 3/15 = 0.2$. As illustrated in Fig.~\ref{fig: safe-pipeline}, we transform sparse pass/fail outcomes into a signal that can be propagated backward through the \OURS \ pipeline. This formal verification procedure utilizes formal methods techniques to \emph{achieve automated labeling and eliminate the need for human annotations}. Additionally, formal verification provides mathematical guarantees to the verified plans, which can be seamlessly adapted to safety-critical applications.

\begin{figure}[t]
    \centering
    \includegraphics[width=\linewidth]{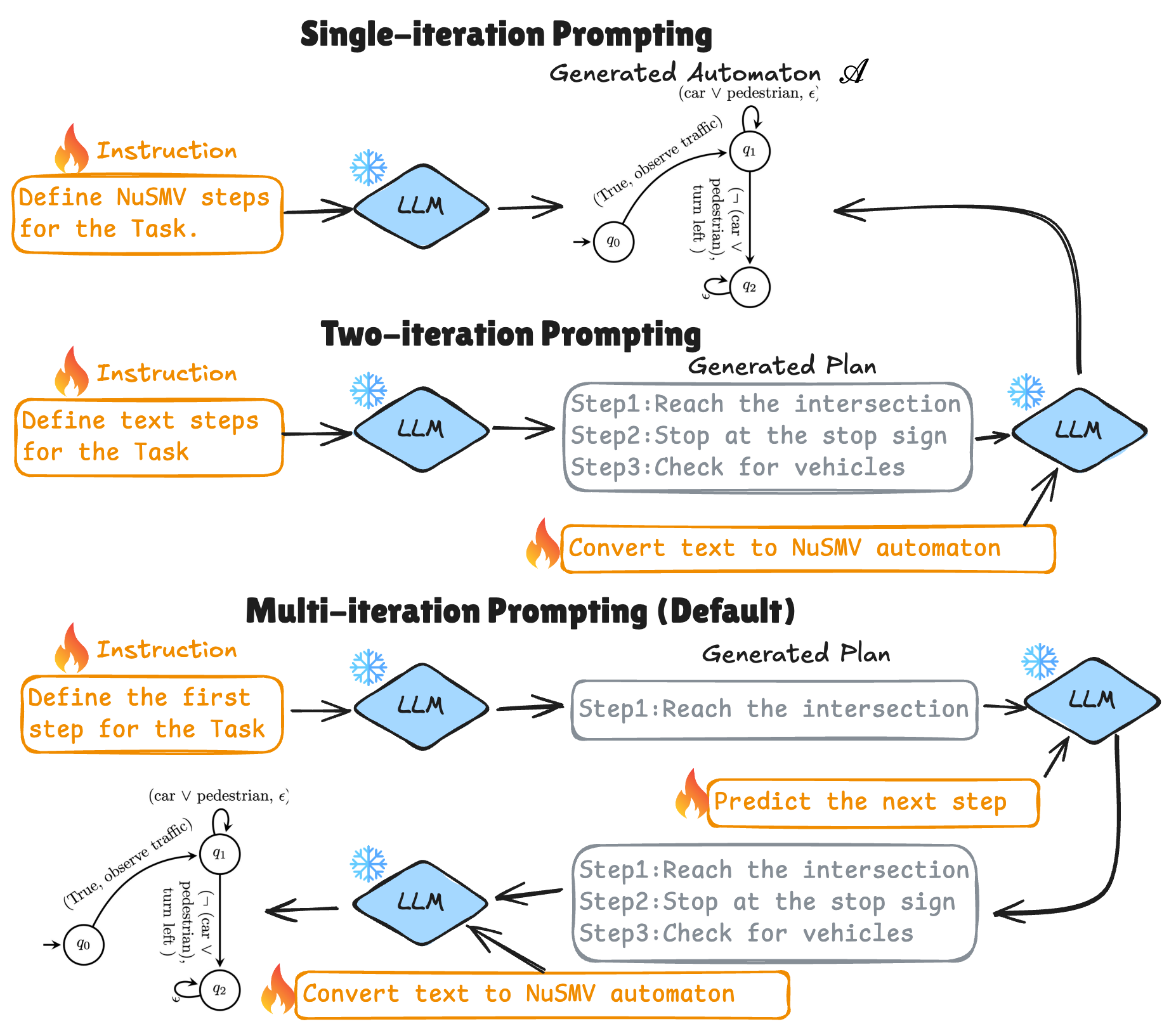}
    \caption{A demonstration of single- and multi-iteration queries. All optimizable prompts are marked in orange boxes.}
    \vspace{-15pt}
    \label{fig: sequence-demo}
\end{figure}

\paragraph{Prompting Strategies.} 
To systematically generate safety-compliant plans, we explore three prompting strategies, as shown in Fig.~\ref{fig: sequence-demo}. \textit{Single-iteration prompting} directly asks the LLM to output a NuSMV-based automaton from the task description. \textit{Two-iteration prompting} first decomposes the task into natural-language step descriptions (e.g., ``reach the intersection,'' ``stop at the stop sign''), which are then converted into an automaton. This strategy adds an extra thinking step for the LLM to generate plans. \textit{Multi-iteration prompting} further decomposes the process into a sequence of partial predictions, where the LLM generates one step at a time and iteratively expands the plan until completion. We employ this multi-iteration prompting strategy by default because it better captures sequential dependencies and aligns with the iterative nature of decision-making, resulting in higher specification compliance in practice.
\section{Experiments}

We evaluate the proposed \OURS \ in safety-constrained robot planning tasks. We demonstrate three claims in the experiments: 
(1) \OURS \ improves the compliance of LLM-generated plans with safety specifications compared to existing prompt optimization baselines.
(2) \OURS \ is more data- and computationally-efficient than fine-tuning approaches while achieving the same level of performance.
(3) \OURS \ maintains both compliance and generalization across varied task settings, specifications, and prompting strategies. 

We benchmark \OURS \ against state-of-the-art approaches and conduct ablation studies to analyze the contribution of different components. We further provide insights into why our approach achieves better performance than related textual gradient methods in sequential decision-making scenarios.

\subsection{Experimental Setup}

\paragraph{Baselines and Evaluation Metric}
We select two current state-of-the-art LLM optimization methods and two prompting methods that apply to safety-constrained planning as benchmarks:

\noindent\textbf{RLVF} \cite{yang-mlsys}: The method first extracts formally verifiable plans from the LLM and verifies the plans. Next, it ranks the plans based on the number of specifications each plan satisfies. Then, it applies DPO \cite{dpo}, which utilizes ranked plans to fine-tune the LLM parameters, ensuring the LLM prefers to generate the higher-ranked plans.

\noindent\textbf{Prompt+Spec}: A simple prompting baseline where the natural language task description and a set of specifications are directly provided to the LLM. Note that the LLM may make mistakes and generate plans that violate the specifications, even when we provide the specifications as inputs.

\noindent\textbf{ICL}: We manually provide a set of input–output examples as in-context demonstrations. The LLM is then queried with a new task description and expected to follow the semantics of the examples.

We include three variants of \OURS:

\noindent \textbf{\OURS (TextGrad)} \cite{yuksekgonul2024textgrad}: The method enables backpropagation of textual feedback to optimize elements, such as prompts or solutions. To adapt this method to our planning tasks, we again use the percentage of specifications being violated as the feedback (e.g., loss).

\noindent \textbf{\OURS (Adalflow)}: The \OURS \ integrates formal verification outcomes into the LLM-AutoDiff pipeline to iteratively optimize prompts for safety compliance. By default, \textbf{Ours} and \textbf{\OURS} \ both refer to \textbf{\OURS (Adalflow)}.

\noindent \textbf{Ours + ICL}: Combines our \OURS (Adalflow) optimization with in-context demonstrations. This hybrid baseline tests whether incorporating demonstrations alongside iterative prompt optimization leads to further improvements in specification compliance.

For evaluation, we define \textit{safety score} $= 1 - n_f/n_{\text{total}}$, where $n_f$ is the number of violated specifications and $n_{\text{total}}$ is the total number of specifications. A higher Safety Score indicates better compliance. In our experiments, we set $n_{\text{total}} = 15$.

\paragraph{Implementation Details}
We use \texttt{GPT-4o-2024} \texttt{-08-16} to generate NuSMV-based plans as the final outcome. We present a sample plan in Fig. \ref{fig: plan-example}.

During evaluation, we generate plans for a \texttt{Jackal ground navigation robot} and propose 15 temporal logic specifications regarding driving safety. For example,
\begin{equation}
    \text{G (Pedestrian }\rightarrow \text{F (Action = Stop))}
    \label{eq: nav_spec}
\end{equation}
means ``\textit{Always (G)} stop \textit{after (F)} a pedestrian is observed.'' 
% We present the complete list of specifications in Listing \ref{lst:safe-specs} in \Cref{sec: safe-prompt-output}. 
The specifications are over the set $AP$ of propositions $AP = \{$ Pedestrian, Opposite Car, Green Light, Green Left Turn Light, Stop Sign, Car From Left, Car From Right, Stop, Move Forward, Turn Left, Turn Right $\}$.

\subsection{Quantitative Evaluation Against Baselines}

\begin{figure}[t]
    \centering
    \includegraphics[width=\linewidth]{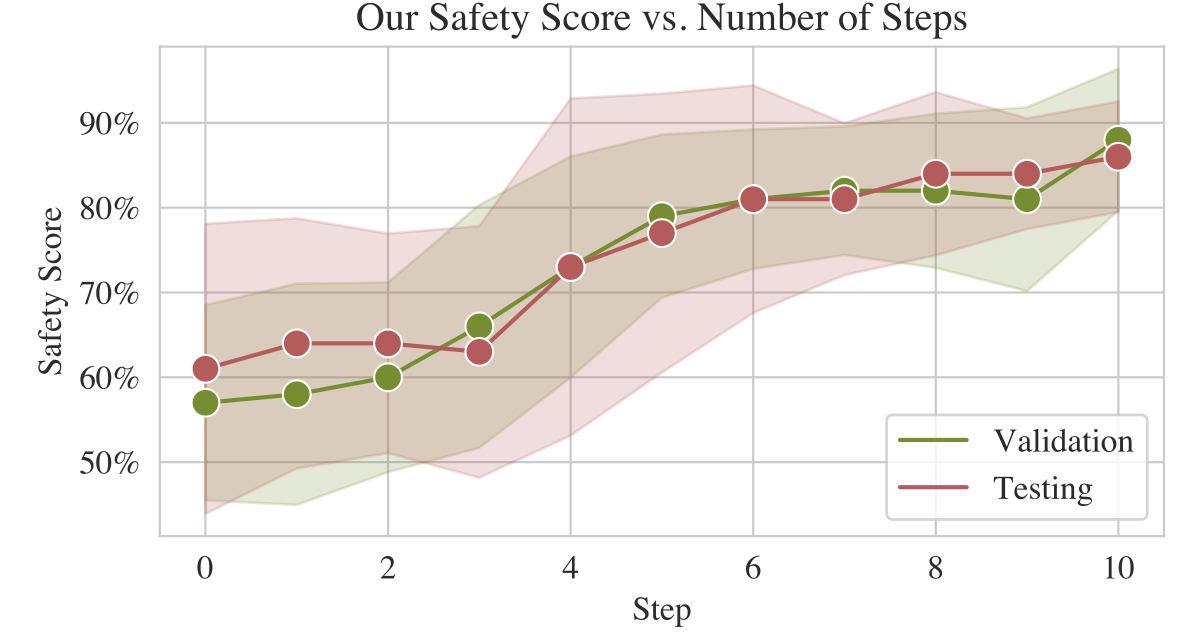}
    \vspace{-15pt}
    \caption{Convergence Comparison. The top figure shows the safety scores achieved by \OURS \ at each re-prompting step. We optimize the input prompts 10 times iteratively, using 20 samples each time to compute the losses. Error bars represent the standard deviations.}
    \label{fig: safety-optimization-step}
\end{figure}

\begin{figure}[t]
    \centering
    \includegraphics[width=\linewidth]{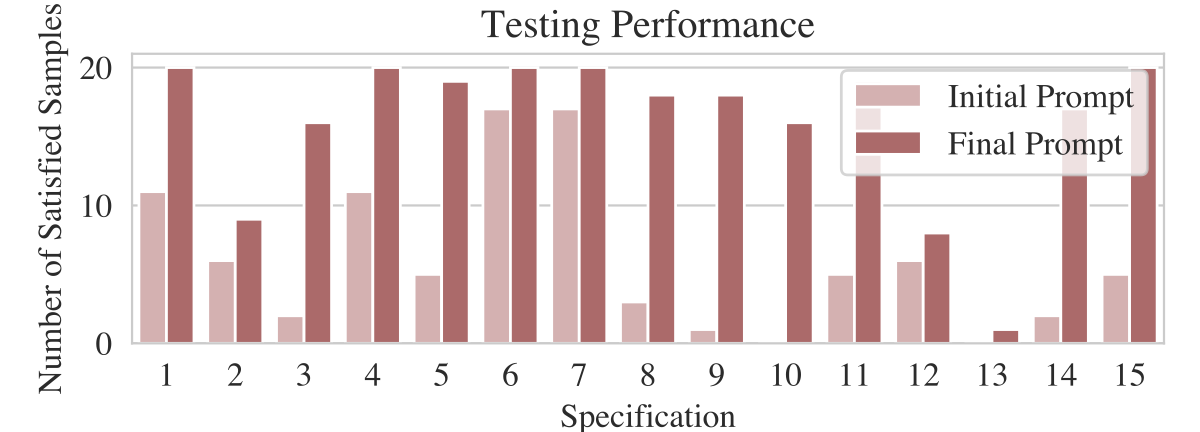}
    \vspace{-0.1in}
    \caption{Specification-level improvements. We examine 20 samples per specification before and after optimization (10 steps). \OURS \ nearly doubles the satisfaction rate across all specifications.}
    \vspace{-15pt}
    \label{fig: safety-test}
\end{figure}

\begin{table}[t]
    \centering
    \begin{tabular}{c|c|c}
    \hline
       \textbf{Method}  & \textbf{Safety Score (Validation)} & \textbf{Safety Score (Test)} \\
       \hline
       RLVF & 0.978 $\pm$ 0.032 & 0.978 $\pm$ 0.032 \\
       \hline
       Prompt+Spec & 0.156 $\pm$ 0.041 & 0.013 $\pm$ 0.013 \\
       ICL & 0.825 $\pm$ 0.015 & 0.800 $\pm$ 0.021 \\
       \OURS (TextGrad) & 0.725 $\pm$ 0.025 & 0.683 $\pm$ 0.024 \\
       \textbf{\OURS} & 0.880 $\pm$ 0.075 & 0.860 $\pm$ 0.058 \\
       \textbf{\OURS + ICL} & \textbf{0.950} $\pm$ 0.036 & \textbf{0.950} $\pm$ 0.072 \\
        \hline
    \end{tabular}
    \caption{Safety score comparison between the baselines. Our method is the best among the training-free methods.}
    \label{tab: safe-score-compare}
\end{table}

We compare \OURS (Adalflow) \ with prompting-based baselines: Prompt+Spec, ICL, and \OURS (TextGrad), and the fine-tuning baseline RLVF. The safety scores of the baselines are summarized in Table~\ref{tab: safe-score-compare}, with convergence behaviors of \OURS (Adalflow) shown in Fig.~\ref{fig: safety-optimization-step} and specification-level improvements in Fig.~\ref{fig: safety-test}.

First, \OURS (Adalflow) \textit{consistently outperforms all prompting-based baselines.} It achieves the highest safety score on both validation and test sets compared with the prompting-based approaches. 

Second, \OURS \ \textit{achieves performance comparable to the fine-tuning baseline with a much faster convergence speed.} As shown in Table~\ref{tab: safe-score-compare}, RLVF achieves the highest score overall, but requires extensive fine-tuning with many epochs. In contrast, \OURS \ reaches a similar performance level without updating model parameters. Moreover, when combined with in-context demonstrations (Ours + ICL), it nearly matches RLVF’s performance while remaining parameter-free.

% On the other hand, Fig.~\ref{fig: safety-optimization-step} (left) shows that \OURS \ improves rapidly within a few re-prompting iterations. At the same time, RLVF (right) requires multiple training epochs to approach similar levels of compliance. This demonstrates that our method is significantly more data- and computation-efficient compared to fine-tuning approaches.

\begin{figure}[t]
    \centering
    \includegraphics[width=0.9\linewidth]{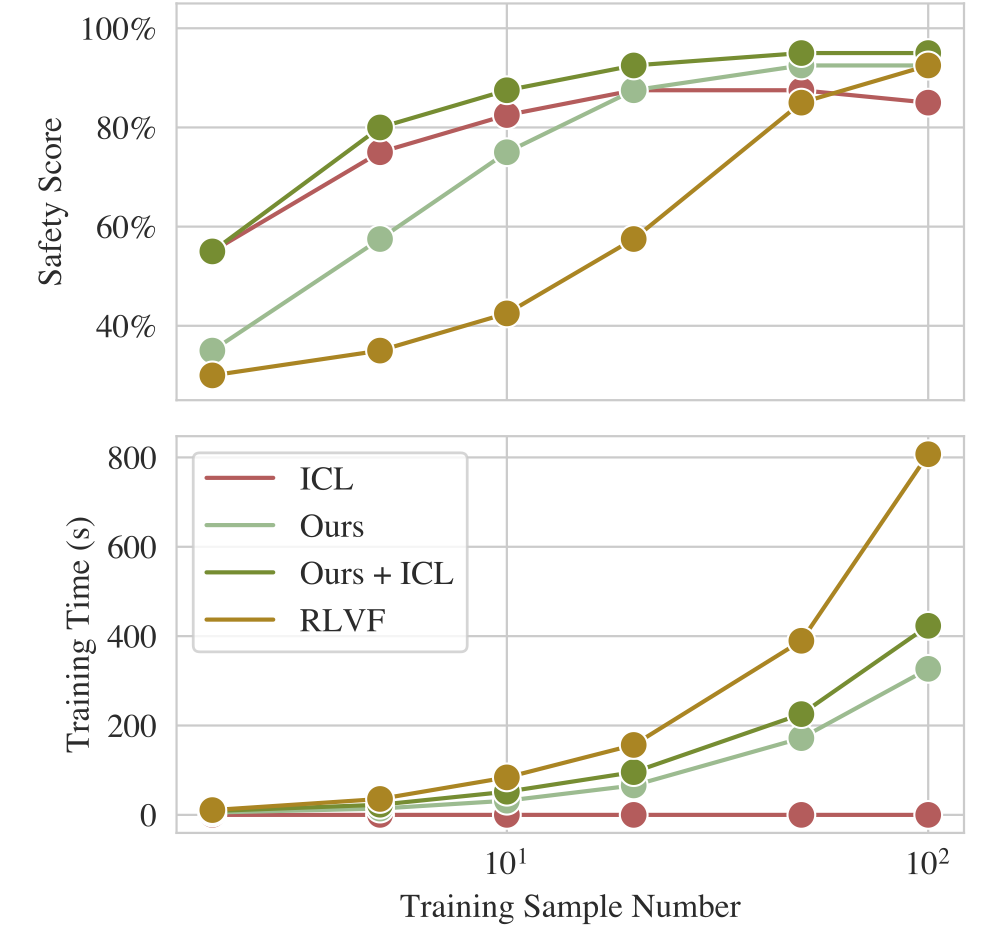}
    \vspace{-0.1in}
    \caption{Safety scores achieved by different methods versus training sample sizes. The figure demonstrates the performance-efficiency trade-offs of \OURS \ and the fine-tuning method. \OURS\ achieves similar safety scores with the fine-tuning method while halving the training time.}
    \label{fig: sample-size}
\end{figure}

\begin{table*}[t]
    \centering
    \begin{tabular}{c|c|c|c||c|c|c||c|c}
    \hline
       \multirow{2}{*}{Step (Validation)}  & \multicolumn{3}{c}{Number of Specs} & \multicolumn{3}{c}{Number of Propositions} & \multicolumn{2}{c}{Optimizer} \\
       \cline{2-9}
       & 3 & 5 & 10 & 5 & 8 & 11 & 4o & o3 \\
       \hline
       0  & 0.400 & 0.490 & 0.470 & 0.400 & 0.270 & 0.120 & 0.400 & 0.400\\
       10 & 0.880 & 0.753 & 0.712 & 0.880 & 0.765 & 0.696 & 0.880 & 0.894\\
       \hline
       Step (Test) & 3 & 5 & 10 & 5 & 8 & 11 & 4o & o3 \\
       \hline
       0  & 0.317 & 0.350 & 0.365 & 0.317 & 0.216 & 0.130 & 0.317 & 0.317\\
       10 & 0.860 & 0.668 & 0.710 & 0.860 & 0.759 & 0.707 & 0.860 & 0.865\\
       \hline
    \end{tabular}
    \caption{Safety scores achieved by our pipeline at different numbers of specifications and propositions. By default, we set the number of specifications to three, the number of propositions to five, and the optimizer to GPT-4o. The table displays \OURS's performance at various complexity levels of specifications and different optimizer models.}
    \label{tab:safety-abalation}
\end{table*}

\subsection{Ablation Studies}
We then conduct ablation studies to test the robustness and efficiency of our approach. Table~\ref{tab:safety-abalation} shows the safety scores achieved by \OURS \ under varying numbers of safety specifications, different specification complexities (measured by the number of propositions), and different optimizers. Across all settings, the prompts optimized by \OURS \ consistently yield significant improvements compared to the unoptimized prompts. 
% Moreover, the safety scores remain stable across variations in specification count, complexity, and optimizer choice, demonstrating the robustness of \OURS.

We also compare the performance of \OURS \ RLVF under different training sample sizes and show the results in Fig.~\ref{fig: sample-size}. \OURS \ \textit{achieves better performance–efficiency trade-off}. While RLVF can reach a high safety score, it requires extensive fine-tuning and large training datasets. In contrast, \OURS \ quickly achieves high safety scores with less than half the number of samples. Hence, combining our optimization with a small number of in-context examples offers a practical and efficient alternative to costly fine-tuning.

\subsection{From Single to Multi-iteration Prompting}

We evaluate different prompting methods and compare \OURS (Adalflow) \ against the \OURS (TextGrad). Figure~\ref{fig: sequence-demo} illustrates the prompting strategies we consider: single-iteration query, two-iteration query, and multi-iteration query. 

\begin{table}[h]
    \centering
    \begin{tabular}{c|c|c}
    \hline
       \textbf{Query Iteration} & \textbf{Optimization Method}   & \textbf{Safety Score (Test)} \\
       \hline
       \multirow{2}{*}{Single} & TextGrad & 0.775 $\pm$ 0.025 \\
        & Adalflow & \textbf{0.805 $\pm$ 0.021} \\
       \hline
       \multirow{2}{*}{Two} & TextGrad & 0.683 $\pm$ 0.024 \\
        & Adalflow & \textbf{0.850 $\pm$ 0.111} \\
       \hline
       \multirow{2}{*}{Multi} & TextGrad & 0.650 $\pm$ 0.041 \\
        & Adalflow & \textbf{0.860 $\pm$ 0.058} \\
       \hline
       % \hline
       % TextGrad & Single & 0.775 $\pm$ 0.025 \\
       % TextGrad & Two & 0.683 $\pm$ 0.024 \\
       % TextGrad & Multi & 0.650 $\pm$ 0.041 \\
       % \hline
       % Adalflow & Single & 0.805 $\pm$ 0.021 \\
       % Adalflow & Two & 0.850 $\pm$ 0.111 \\
       % Adalflow & Multi & \textbf{0.883 $\pm$ 0.024} \\
       % \hline
    \end{tabular}
    \caption{Comparison between \OURS (Adalflow) and (TextGrad) under different query methods. \OURS (Adalflow) achieves higher safety scores as the query iteration increases, whereas TextGrad's scores are degraded. 
    }
    \label{tab: sequence-compare}
\end{table}

Table~\ref{tab: sequence-compare} presents the safety scores and the average response times under these prompting methods. In the result, since Adalflow can handle sequential prompting, \OURS (Adalflow) consistently outperforms \OURS (TextGrad) across all prompting strategies. In particular, \textit{the advantage of using Adalflow backbone is most pronounced in the multi-iteration query}. In contrast, TextGrad is less effective as the decision-making task involves more iterations. In addition, the result also indicates the \emph{compatibility of \OURS \ to varied query formats}, i.e., no re-training is needed for different query formats.

\section{Generalization to Real Robots}

To assess the practicality of our approach, we deploy \OURS \ in real-world robotic settings. Through real robot deployments, we demonstrate that the prompts optimized by \OURS \ can guide the LLM to produce specification-compliant plans for robot execution.

The demonstrations indicate that \OURS \ successfully generalizes to real-robot deployments. In navigation tasks such as “go straight at the intersection,” “turn left safely,” or “navigate to the lounge while avoiding pedestrians,” the optimized prompts yield executable plans that satisfy safety constraints during execution.

\begin{figure*}[t]
    \centering
    \includegraphics[width=0.8\linewidth]{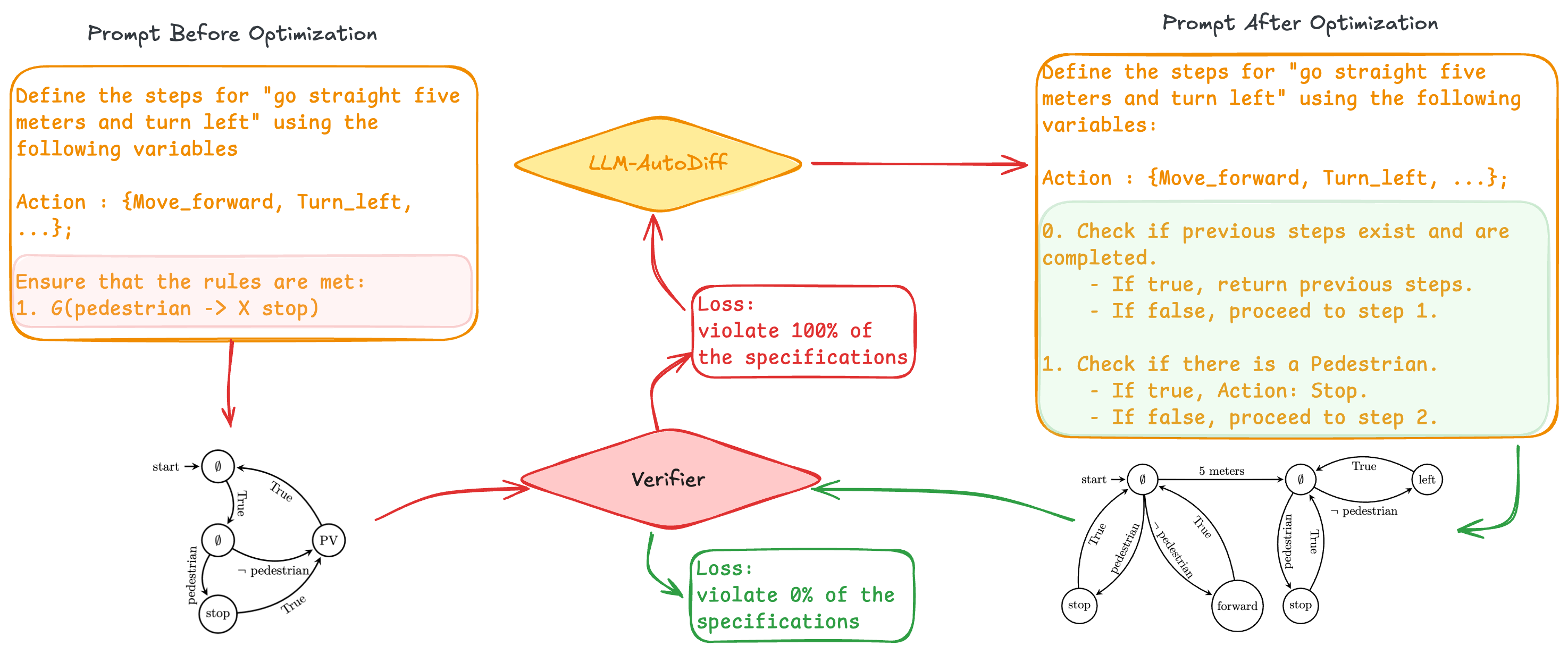}
    \caption{A step-by-step illustration of prompt optimization on robot navigation.}
    \label{fig: prompt-example}
\end{figure*}

\begin{figure*}[t]
    \centering
    \includegraphics[width=0.49\linewidth]{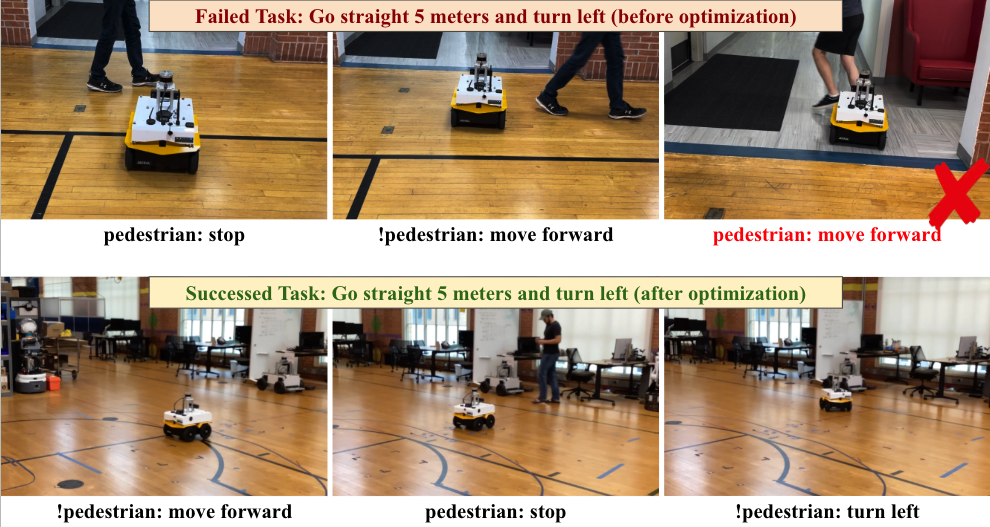}
    \includegraphics[width=0.49\linewidth]{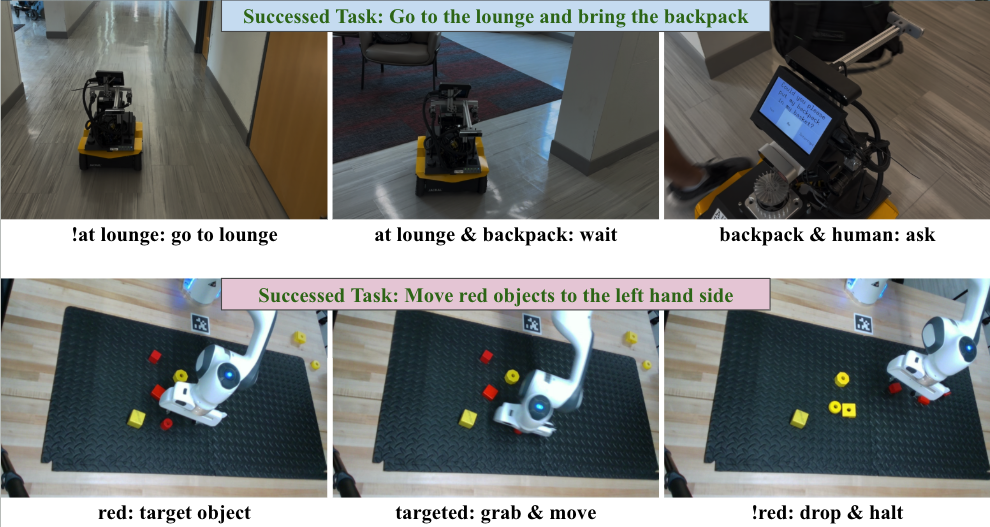}
    \caption{Demonstrations of real-robot deployment. We deploy \OURS \ on a Jackal Clearpath robot (left), a Jackal indoor robot (top right), and a robot arm (bottom right) to complete navigation, delivery, and table-top manipulation tasks. Optimized prompts yield plans that transfer successfully to real execution, reducing safety violations compared to unoptimized prompting.}
    \label{fig: robot-demo}
\end{figure*}

\begin{figure}[t]
    \vspace{-0.2in}
    \centering
    \includegraphics[width=0.85\linewidth]{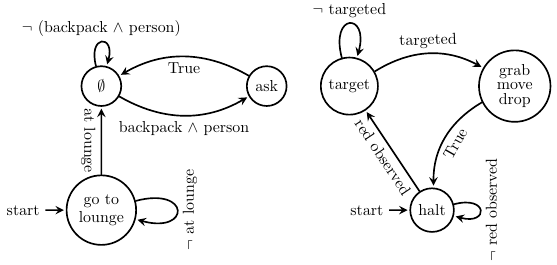}
    \caption{The left and right automata represent the plans for robot delivery (top right of Fig. \ref{fig: robot-demo}) and table-top manipulation (bottom right of Fig. \ref{fig: robot-demo}) tasks.}
    \vspace{-15pt}
    \label{fig: automaton}
\end{figure}

\subsection{Robot Demonstration}
\label{sec: real-robot-demo}
We deploy a \emph{Jackal Clearpath robot} to perform navigation tasks and demonstrate, step by step, how \OURS \ generates verifiable and executable plans in Fig. \ref{fig: prompt-example}. For clarity of this demonstration, we focus on a single representative safety specification, shown in Equation~\ref{eq: nav_spec}.

First, the automaton generated via the initial prompt (bottom left in Fig. \ref{fig: prompt-example}) fails the specification in \ref{eq: nav_spec}. The pipeline obtains feedback from the verifier and formulates a textual loss, which is then used to optimize the prompt. Next, we feed the optimized prompt into the LLM and obtain an automaton as presented in Fig. \ref{fig: prompt-example} (bottom left). This automaton satisfies the specification, yielding zero violations. We then deploy the verified plan in the real environment, as illustrated in Fig.~\ref{fig: robot-demo}, demonstrating that the robot executes the task safely and in full compliance with the specification.

Besides the improvement in specification compliance, we also demonstrate better interpretability compared with ordinary gradient update methods, as shown in Fig. \ref{fig: prompt-example}. \OURS \ provides human-readable loss for refining prompts, which enables the tracing of how verification feedback results in concrete modifications to task instructions. 

\subsection{Out-of-Domain Generalization}

\begin{table}[t]
    \centering
    \begin{tabular}{c|c|c|c}
    \hline
       \textbf{Method}  & \textbf{Jackal Clearpath} & \textbf{Jackal Indoor} & \textbf{Robot Arm} \\
       \hline
       Prompt+Spec & 0.45 & 0.43 & 0.60 \\
       ICL & 0.75 & 0.83 {\scriptsize +0.08} & 0.80 {\scriptsize +0.05}  \\
       RLVF & 0.90 & 0.63 {\scriptsize -0.27} & 0.75 {\scriptsize -0.15} \\
       \textbf{\OURS} & \textbf{0.90} & \textbf{0.88} {\scriptsize -0.02} & \textbf{0.85} {\scriptsize -0.05} \\
        \hline
    \end{tabular}
    \caption{Safety scores across different robotic domains. 
    % Optimized prompts trained on navigation tasks (Jackal Clearpath) generalize to indoor delivery and robot arm manipulation tasks without re-optimization. 
    We display the generalization gap alongside the safety score. \texttt{Prompt+Spec} provides a baseline safety score without any optimization. \texttt{ICL} requires human-provided examples in the prompt and \texttt{RLVF}'s safety scores drop significantly. Our \texttt{\OURS} maintains high safety scores in out-of-domain tasks without any human interference.  
    }
    \vspace{-15pt}
    \label{tab: out-domain-test}
\end{table}

We test whether prompts optimized for navigation (Jackal Clearpath) can be generalized to other robotic domains. Specifically, we evaluate an indoor delivery task with a \emph{Jackal indoor robot} and a table-top manipulation task with a \emph{robot arm}. For these experiments, only the propositions and specifications are redefined. At the same time, the structures and wordings of the optimized prompts remain the same as the prompts presented in Sec. \ref{sec: real-robot-demo}.

The propositions $AP$ and specifications $\Phi$ for the Jackal indoor robot and the robot arm are

$AP_{indoor} = \{$at lounge, at classroom, backpack observed, human observed, go to lounge, go to classroom, ask, wait$\}$,

$AP_{arm} = \{$red block observed, block targeted, target object, grab, move, drop, halt$\}$,

$\Phi_{indoor} = \{$G (human observed $\rightarrow$ F ask), \\ G (at lounge \& !human observed $\rightarrow$ X wait )$ \}$,

$\Phi_{arm} = \{$G(! red $\rightarrow \neg$ ! X grab )$\}$,

where X, F, G means ``next,'' ``eventually,'' and ``always.''

For visual demonstration, we select a task ``go to the lounge and bring the backpack'' for the indoor robot and a task ``move red objects to the left-hand side'' for the robot arm. We present the visual representations of the generated automaton-based plans (in NuSMV) in Fig. \ref{fig: automaton}. The automata for both tasks passed the verification step and were successfully executed in the real environment. We show the execution recordings for both tasks in Fig. \ref{fig: robot-demo}.

Quantitatively, Table~\ref {tab: out-domain-test} shows the safety scores across various robotic tasks. We query the plans for 20 tasks per robotic domain and compute the average safety scores. \OURS \ maintains consistent safety scores across all domains, demonstrating that the optimized prompts capture general safety reasoning patterns rather than overfitting to a specific robot or task. While ICL requires human-provided in-context examples for new domains, \OURS \ generalizes to these new domains without human in the loop. Simultaneously, \OURS \ outperforms the RLVF fine-tuning baseline on the out-of-domain tasks, showcasing better generalizability compared with the fine-tuning methods.

Notably, we show that \textit{optimized prompts trained on navigation tasks can also be applied to other robotic tasks, such as robot arm manipulation, without re-optimization}. By only re-specifying the constraints and task descriptions in the optimized prompt format, the LLM can generate plans that meet the new constraints. This demonstrates that the improvements obtained through \OURS  \ are not task-specific but generalize across domains, further underscoring the scalability of our approach.

\section{Conclusion}

We introduced \OURS, a fine-tuning-free framework that combines prompt optimization with formal verification feedback to align language models with safety specifications. Empirical results indicate that \OURS \ outperforms prompting-based baselines, achieves compliance comparable to fine-tuning methods with far greater efficiency, and generalizes across different tasks and robot platforms without re-optimization. By treating prompts as trainable parameters, our approach enables transparent and auditable improvements, paving the way for scalable and trustworthy LLM-driven control. In future work, we plan to extend \OURS \ to multimodal inputs such as vision and language, explore probabilistic guarantees for specification satisfaction, and investigate its application to broader domains where safety and verifiability are critical, such as medical applications.

\section*{Acknowledgement}
This work was supported in part by ONR under Grant No. N00014-25-1-2479; by ARL under Grant No. W911NF-23-S-0001; by DARPA under Grant No. HR0011-24-9-0431 and RTX CW2231110; and by the São Paulo Research Foundation (FAPESP) under grant 2022/15304-4.

\bibliographystyle{IEEEtran}
\bibliography{references}

@article{pryzant2023automatic,
  title   = {Automatic prompt optimization with" gradient descent" and beam search},
  author  = {Pryzant, Reid and Iter, Dan and Li, Jerry and Lee, Yin Tat and Zhu, Chenguang and Zeng, Michael},
  journal = {arXiv preprint arXiv:2305.03495},
  year    = {2023}
}

@article{yuksekgonul2024textgrad,
  title   = {TextGrad: Automatic" Differentiation" via Text},
  author  = {Yuksekgonul, Mert and Bianchi, Federico and Boen, Joseph and Liu, Sheng and Huang, Zhi and Guestrin, Carlos and Zou, James},
  journal = {arXiv preprint arXiv:2406.07496},
  year    = {2024}
}

@article{brown2020language,
  title   = {Language models are few-shot learners},
  author  = {Brown, Tom B},
  journal = {arXiv preprint arXiv:2005.14165},
  year    = {2020}
}

@article{yao2022react,
  title   = {React: Synergizing reasoning and acting in language models},
  author  = {Yao, Shunyu and Zhao, Jeffrey and Yu, Dian and Du, Nan and Shafran, Izhak and Narasimhan, Karthik and Cao, Yuan},
  journal = {arXiv preprint arXiv:2210.03629},
  year    = {2022}
}

@article{zhou2022large,
  title   = {Large language models are human-level prompt engineers},
  author  = {Zhou, Yongchao and Muresanu, Andrei Ioan and Han, Ziwen and Paster, Keiran and Pitis, Silviu and Chan, Harris and Ba, Jimmy},
  journal = {arXiv preprint arXiv:2211.01910},
  year    = {2022}
}

@article{suzgun2022challenging,
  title   = {Challenging big-bench tasks and whether chain-of-thought can solve them},
  author  = {Suzgun, Mirac and Scales, Nathan and Sch{\"a}rli, Nathanael and Gehrmann, Sebastian and Tay, Yi and Chung, Hyung Won and Chowdhery, Aakanksha and Le, Quoc V and Chi, Ed H and Zhou, Denny and others},
  journal = {arXiv preprint arXiv:2210.09261},
  year    = {2022}
}

@article{achiam2023gpt,
  title   = {Gpt-4 technical report},
  author  = {Achiam, Josh et al.},
  journal = {arXiv preprint arXiv:2303.08774},
  year    = {2023}
}

@book{Clarke2018modelchecking,
  author       = {Edmund M. Clarke and
                  Orna Grumberg and
                  Daniel Kroening and
                  Doron A. Peled and
                  Helmut Veith},
  title        = {Model checking, 2nd Edition},
  publisher    = {{MIT} Press},
  address      = {Cambridge, Massachusetts, USA},
  year         = {2018}
}

@article{yang2024joint,
  title={Joint Verification and Refinement of Language Models for Safety-Constrained Planning},
  author={Yang, Yunhao and Ward, William and Hu, Zichao and Biswas, Joydeep and Topcu, Ufuk},
  journal={arXiv preprint arXiv:2410.14865},
  year={2024}
}

@inproceedings{Cimatti2002NuSMV,
  author    = {Alessandro Cimatti et al.},
  title     = {Nu{SMV} 2: An OpenSource Tool for Symbolic Model Checking},
  booktitle = {Computer Aided Verification},
  series    = {Lecture Notes in Computer Science},
  publisher = {Springer},
  address = {NY, USA},
  volume    = {2404},
  pages     = {359--364},
  year      = {2002}
}

@inproceedings{yang-mlsys,
  author       = {Yunhao Yang and
                  Neel P. Bhatt et al.},
  title        = {Fine-Tuning Language Models Using Formal Methods Feedback: {A} Use
                  Case in Autonomous Systems},
  booktitle    = {Conference on Machine Learning and
                  Systems},
  publisher    = {mlsys.org},
  address = {CA, USA},
  year         = {2024}
}

@inproceedings{yang2024aamas,
  author       = {Yunhao Yang and
                  Cyrus Neary and
                  Ufuk Topcu},
  title        = {Multimodal Pretrained Models for Verifiable Sequential Decision-Making:
                  Planning, Grounding, and Perception},
  booktitle    = {International Conference on Autonomous Agents and Multiagent Systems},
  address = {New Zealand},
  pages        = {2011--2019},
  publisher    = {ACM},
  year         = {2024}
}

@article{dpo,
  title={Direct preference optimization: Your language model is secretly a reward model},
  author={Rafailov, Rafael and Sharma, Archit and Mitchell, Eric and Manning, Christopher D and Ermon, Stefano and Finn, Chelsea},
  journal={Advances in Neural Information Processing Systems},
  volume={36},
  pages={53728--53741},
  year={2023}
}

@article{yin2025llm-autodiff,
  title={LLM-AutoDiff: Auto-Differentiate Any LLM Workflow},
  author={Yin, Li and Wang, Zhangyang},
  journal={arXiv preprint arXiv:2501.16673},
  year={2025}
}

@article{sordoni2023joint,
  title={Joint prompt optimization of stacked llms using variational inference},
  author={Sordoni, Alessandro and et al.},
  journal={Advances in Neural Information Processing Systems},
  volume={36},
  pages={58128--58151},
  year={2023}
}

@inproceedings{yang2024large,
  title={Large Language Models as Optimizers},
  author={Yang, Chengrun and Wang, Xuezhi and Lu, Yifeng and Liu, Hanxiao and Le, Quoc V and Zhou, Denny and Chen, Xinyun},
  booktitle={The Twelfth International Conference on Learning Representations},
year={2024}
}

@inproceedings{khattab2024dspy,
  title={Dspy: Compiling declarative language model calls into state-of-the-art pipelines},
  author={Khattab, Omar and Singhvi, Arnav and others},
  booktitle={The Twelfth International Conference on Learning Representations},
  year={2024}
}

@inproceedings{wangpromptagent,
  title={PromptAgent: Strategic Planning with Language Models Enables Expert-level Prompt Optimization},
  author={Wang, Xinyuan and Li, Chenxi and Wang, Zhen and Bai, Fan and Luo, Haotian and Zhang, Jiayou and Jojic, Nebojsa and Xing, Eric and Hu, Zhiting},
  booktitle={The Twelfth International Conference on Learning Representations},
  year={2024}
}

@article{agrawal2025gepa,
  title={Gepa: Reflective prompt evolution can outperform reinforcement learning},
  author={Agrawal, Lakshya A et al.},
  journal={arXiv preprint arXiv:2507.19457},
  year={2025}
}

@inproceedings{Ouyang2022FollowInstructions,
  author       = {Long Ouyang et al.},
  title        = {Training Language Models to Follow Instructions with Human Feedback},
  booktitle    = {Advances in Neural Information Processing Systems},
  address = {New Orleans, LA, USA},
  year         = {2022}
}

@article{RLHF,
  author       = {Nisan Stiennon and
                  Long Ouyang and
                  Jeff Wu and
                  Daniel M. Ziegler and
                  Ryan Lowe and
                  Chelsea Voss and
                  Alec Radford and
                  Dario Amodei and
                  Paul F. Christiano},
  title        = {Learning to summarize from human feedback},
  journal      = {arXiv preprint arXiv:2009.01325},
  year         = {2020}
}

@article{Rafailov2023DPO,
  author       = {Rafael Rafailov and
                  Archit Sharma and
                  Eric Mitchell and
                  Stefano Ermon and
                  Christopher D. Manning and
                  Chelsea Finn},
  title        = {Direct Preference Optimization: Your Language Model is Secretly a Reward Model},
  journal      = {arXiv preprint arXiv:2305.18290},
  year         = {2023}
}

@inproceedings{bhattknow,
  title={Know Where You’re Uncertain When Planning with Multimodal Foundation Models: A Formal Framework},
  author={Bhatt, Neel P and Yang, Yunhao and Siva, Rohan and Milan, Daniel and Wang, Zhangyang and Topcu, Ufuk},
  booktitle={Eighth Conference on Machine Learning and Systems},
  address = {Santa Clara, CA, USA},
  year = {2025}
}

@misc{Song2022LLMPlannerFG,
  title={LLM-Planner: Few-Shot Grounded Planning for Embodied Agents with Large Language Models},
  author={Chan Hee Song and Jiaman Wu and Clay Washington and Brian M. Sadler and Wei-Lun Chao and Yu Su},
  journal={ArXiv preprint arXiv:2212.04088},
  year={2022}
}

@misc{Singh2022ProgPromptGS,
  title={ProgPrompt: Generating Situated Robot Task Plans using Large Language Models},
  author={Ishika Singh and Valts Blukis and Arsalan Mousavian and Ankit Goyal and Danfei Xu and Jonathan Tremblay and Dieter Fox and Jesse Thomason and Animesh Garg},
  journal={ArXiv preprint arXiv:2209.11302},
  year={2022}
}

@article{CodeBotler,
  author       = {Zichao Hu and
                  Francesca Lucchetti and
                  Claire Schlesinger and
                  Yash Saxena and
                  Anders Freeman and
                  Sadanand Modak and
                  Arjun Guha and
                  Joydeep Biswas},
  title        = {Deploying and Evaluating LLMs to Program Service Mobile Robots},
  journal      = {{IEEE} Robotics Autom. Lett.},
  volume       = {9},
  number       = {3},
  pages        = {2853--2860},
  year         = {2024}
}

@misc{Liu2023LLMPEL,
  title={LLM+P: Empowering Large Language Models with Optimal Planning Proficiency},
  author={B. Liu et al.},
  journal={ArXiv preprint arXiv:2304.11477},
  year={2023}
}

@inproceedings{wang2024llm,
  title={Llm-based robot task planning with exceptional handling for general purpose service robots},
  author={Wang, Ruoyu and Yang, Zhipeng and Zhao, Zinan and Tong, Xinyan and Hong, Zhi and Qian, Kun},
  booktitle={2024 43rd Chinese Control Conference (CCC)},
  pages={4439--4444},
  year={2024},
  organization={IEEE}
}

@inproceedings{yang2023planning,
  title={On the planning, search, and memorization capabilities of large language models},
  author={Yang, Yunhao and Tomar, Anshul},
  booktitle={International Conference on Intelligent Vision and Computing},
  pages={24--38},
  year={2023},
  organization={Springer}
}

@article{chen2024can,
  title={Can We Rely on LLM Agents to Draft Long-Horizon Plans? Let's Take TravelPlanner as an Example},
  author={Chen, Yanan and Pesaranghader, Ali and Sadhu, Tanmana and Yi, Dong Hoon},
  journal={arXiv preprint arXiv:2408.06318},
  year={2024}
}

@article{white2023prompt,
  title={A prompt pattern catalog to enhance prompt engineering with ChatGPT},
  author={White, Jules and et al.},
  journal={arXiv preprint arXiv:2302.11382},
  year={2023}
}

@inproceedings{marvin2023prompt,
  title={Prompt engineering in large language models},
  author={Marvin, Ggaliwango and Hellen, Nakayiza and Jjingo, Daudi and Nakatumba-Nabende, Joyce},
  booktitle={International conference on data intelligence and cognitive informatics},
  pages={387--402},
  year={2023},
  organization={Springer}
}

@article{dong2022survey,
  title={A survey on in-context learning},
  author={Dong, Qingxiu and Li, Lei and Dai, Damai and Zheng, Ce and Ma, Jingyuan and Li, Rui and Xia, Heming and Xu, Jingjing and Wu, Zhiyong and Liu, Tianyu and others},
  journal={arXiv preprint arXiv:2301.00234},
  year={2022}
}

\end{document}